\documentclass[sigconf]{acmart}
\usepackage[caption=false,font=normalsize,labelfont=sf,textfont=sf]{subfig}
\usepackage[ruled,linesnumbered]{algorithm2e}
\usepackage{multirow}
\usepackage{microtype}      
\usepackage{boldline}
\usepackage{balance}
\AtBeginDocument{%
  }

\copyrightyear{2023} 
\acmYear{2023} 
\setcopyright{acmlicensed}
\acmConference[MM '23] {Proceedings of the 31st ACM International Conference on Multimedia}{October 29--November 3, 2023}{Ottawa, ON, Canada.}
\acmBooktitle{Proceedings of the 31st ACM International Conference on Multimedia (MM '23), October 29--November 3, 2023, Ottawa, ON, Canada}
\acmPrice{15.00}
\acmISBN{979-8-4007-0108-5/23/10} 
\acmDOI{10.1145/3581783.3611906}

\usepackage{hyperref}
\hypersetup{
	colorlinks=true,
	linkcolor=cyan,
	filecolor=blue,      
	urlcolor=red,
	citecolor=green,
}
\begin{document}

\title{Semi-supervised Semantic Segmentation \\ with Mutual Knowledge Distillation}

\author{Jianlong Yuan}
\affiliation{%
  \institution{Alibaba Group}
  \city{Beijing}
  \country{China}
  }
\email{gongyuan.yjl@alibaba-inc.com}

\author{Jinchao Ge}
\affiliation{%
  \institution{The University of Adelaide}
  \city{Adelaide}
  \country{Australia}
}
\email{jinchao.ge@adelaide.edu.au}

\author{Zhibin Wang}
\affiliation{%
 \institution{zhibin.waz@alibaba-inc.com}
 \city{Hangzhou}
 \country{China}}
\email{zhibin.waz@alibaba-inc.com}

\author{Yifan Liu}
\affiliation{%
  \institution{The University of Adelaide}
  \city{Adelaide}
  \country{Australia}}
\email{yifan.liu04@adelaide.edu.au}

\renewcommand{\shortauthors}{Jianlong Yuan, Jinchao Ge, Zhibin Wang, \& Yifan Liu}

\begin{abstract}
Consistency regularization has been widely studied in recent semi-supervised semantic segmentation methods, and promising performance has been achieved. In this work, we propose a new consistency regularization framework, termed mutual knowledge distillation (MKD), combined with data and feature augmentation. We introduce two auxiliary mean-teacher models based on consistency regularization. More specifically, we use the pseudo-labels generated by a mean teacher to supervise the student network to achieve a mutual knowledge distillation between the two branches. In addition to using image-level strong and weak augmentation, we also discuss feature augmentation. This involves considering various sources of knowledge to distill the student network. Thus, we can significantly increase the diversity of the training samples. Experiments on public benchmarks show that our framework outperforms previous state-of-the-art (SOTA) methods under various semi-supervised settings. Code is available at \href{https://github.com/jianlong-yuan/semi-mmseg}{\textcolor[rgb]{0.88,0.0078,0.52}{semi-mmseg}}.
\end{abstract}



\maketitle

\section{Introduction}
Segmentation is a fundamental task in visual understanding that aims to classify each pixel in an image into a predefined set of categories. While recent works in semantic segmentation ~\cite{zhao2017pyramid,chen2018encoder,wang2020deep,yuan2020multi,xie2021segformer,zhang2021knet} have made significant progress using supervised learning with the use of large-scale datasets\cite{everingham2015pascal,cordts2016cityscapes,mottaghi_cvpr14,zhou2017scene}. However, labeling such datasets can be labor-intensive and time-consuming for dense prediction problems, requiring up to 60 times more effort than image-level labeling~\cite{lin2014microsoft}. To address this limitation, semi-supervised learning~\cite{berthelot2019mixmatch, sohn2020fixmatch, yuan2021simple, yun2019cutmix} attempts to learn a model with a limited set of labeled images and a large set of unlabeled images. 

State-of-the-art semi-supervised semantic segmentation methods employ consistency regularization to enhance the similarity between the outputs of teacher and student during training. Data augmentations are commonly implemented on images, a practice evidenced by studies such as \citep{yuan2021simple,yang2021st++,yun2019cutmix}. Furthermore, they can also be applied to features, as indicated in \cite{ouali2020semi}. Additionally, the utilization of various networks, each possessing distinct initialization parameters, is a common practice, as delineated in \cite{chen2021semi, ke2020guided}. For example, the CPS~\cite{chen2021semi} method feeds the same image into two different initialized networks and uses the pseudo labels generated from one branch to supervise the other branch. However, this method does not preserve important historical information during training. Note that the two branches are optimized with back-propagation without moving average during training. Thus, the model `forgets' important historical information along with the training steps as stated in previous research ~\cite{tarvainen2017mean, he2020momentum, grill2020bootstrap}.

To further improve the performance of the semi-supervised semantic segmentation models, we propose a novel mutual knowledge distillation framework. This framework employs two branches of co-training~\cite{chen2021semi} with different initialized parameters and two auxiliary mean teacher models to record the information during the training process and provide extra supervision. The pseudo labels generated from one teacher network supervise the other student and vice versa. Weak augmentation is applied to teacher input images to increase prediction confidence, while input images from the student networks are strongly augmented to diversify samples. Pseudo labels from the teacher network tend to be more reliable, while the student network can be trained on more diverse and challenging samples.
We explore feature-level augmentation in student networks, drawing inspiration from the implicit semantic data augmentation technique applied in~\cite{NIPS2019_9426, wang2021regularizing}.

\begin{figure*}
\centering
\includegraphics[width=\textwidth]{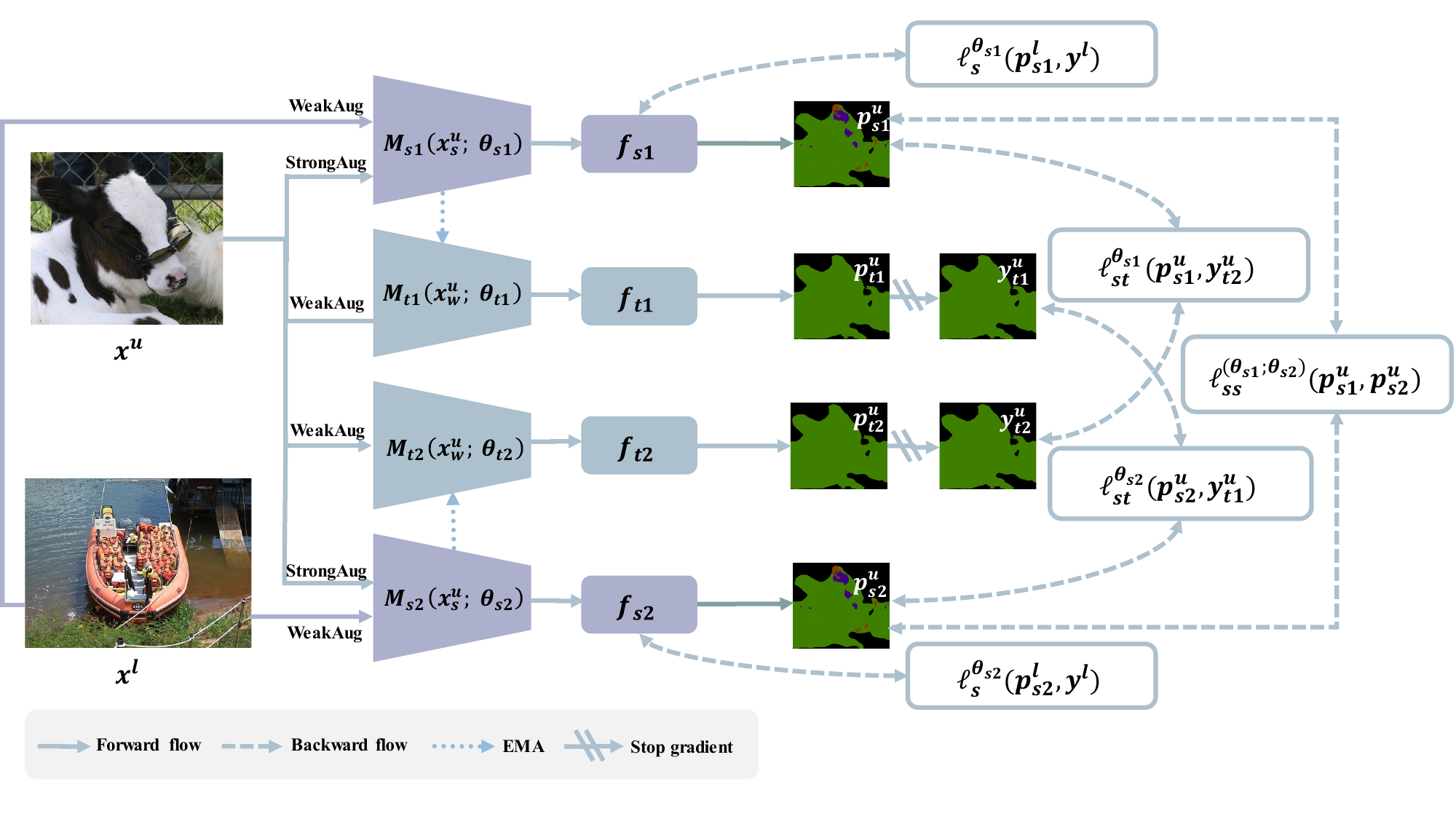}

\caption{For each image $x^u$, we apply weak augmentation (WeakAug) on the teacher network and strong augmentation (StrongAug) on the student network. Here $\mathbf{p}$ denotes the logits. $\theta$ is the parameters of the model and $\mathbf{y}^u$ is the one-hot labels generated from $\mathbf{p}^u$. We train the model by minimizing the consistency loss $\ell_{st}$ and $\ell_{ss}$ on the unlabeled set and the cross-entropy loss $\ell_{s}$ on the labeled set.}

\label{fig: framework}
\end{figure*}

Our approach achieves state-of-the-art performance on the PASCAL VOC 2012~\cite{everingham2015pascal}, Cityscapes~\cite{cordts2016cityscapes}, and COCO~\cite{lin2014microsoft} datasets, under various splits of semi-supervised settings. Our main contributions are summarized as follows:

1), We propose the mutual knowledge distillation framework, a new consistency regularization approach for semi-supervised semantic segmentation($S^4$). This framework involves two different initialized student networks and two corresponding mean teacher networks.  The knowledge from one teacher network is used to supervise the other student branch, and vice versa.

2), We investigate the efficacy of different data augmentation methods for $S^4$. Specifically, we discuss feature-level to enhance the diversity of the training dataset. Furthermore, we employ strong and weak augmentations to the student and teacher networks, respectively.

3), We empirically demonstrate the effectiveness of our approach, which achieves state-of-the-art performance on PASCAL VOC 2012, Cityscapes, and COCO datasets under various semi-supervised settings. A detailed ablation study verifies the usefulness of each component in the proposed framework.
\section{Related Work}
\subsection{Semantic segmentation}
A variety of methods have been proposed for this task~\citep{chen2018encoder,wang2020deep,yuan2020multi,xie2021segformer,zhao2017pyramid,zhang2021knet}, starting with the fully convolutional network (FCN)\cite{long2015fully}, which trains a pixel-level classifier. It is worth mentioning that our work is based on DeepLabV3Plus~\cite{chen2018encoder}, which applies a spatial pyramid pooling structure and an encoder-decoder structure to refine object boundaries. The majority of existing approaches in the literature operate under the fully-supervised regime, wherein a significant amount of labeled data is necessary.

\subsection{Semi-supervised semantic segmentation}
The task of semantic segmentation usually requires a large amount of annotated data, but semi-supervised learning (SSL) offers a new perspective to tackle this problem. SSL methods can be broadly classified into two categories: consistency regularization-based techniques~\cite{tarvainen2017mean, berthelot2019remixmatch, sohn2020fixmatch, chen2021semi, hu2021semi, french2019semi, wang2022semi, souly2017semi, hung2018adversarial}, and self-training based methods~\cite{yuan2021simple, he2021re, yang2021st++, zou2020pseudoseg, xie2020self, chen2020naive}.

Consistency-based methods enforce the model to generate the same prediction from augmented images and original ones. Temporal ensembling~\cite{laine2016temporal} implements the idea of ensemble multiple checkpoints of students. In particular, mean teacher~\cite{tarvainen2017mean, berthelot2019remixmatch, wang2020deep} employs the exponential moving average of the model parameters to update the weights of the teacher model. Moreover, the student model is supervised under the pseudo-label generated by the teacher model.

Co-training for consistency~\cite{ke2020guided, chen2021semi} feeds the same image into two different initialized networks and uses the pseudo labels generated from one branch to supervise the other branch. U$^2$PL\cite{wang2022semi} designs a method to select reliable annotations from unreliable candidate pixels. Self-training ~\cite{yuan2021simple, yang2021st++, xie2020self, he2021re} based methods aim at generating pseudo labels to enlarge the training set. They use a teacher model to generate pseudo-labels based on suitable data augmentation and thresholds. Unlike these methods, our approach designs two student networks and two auxiliary mean teacher networks. Furthermore, we apply augmentation to the images and feature augmentation in the same framework, boosting the student network's performance.

\subsection{Data augmentation}
We describe data augmentation in SSL from two different perspectives: image-level augmentation~\cite{zhang2021robust} and feature-level augmentation~\cite{ge2020mutual, feng2022dmt}. For instance, FixMatch\cite{sohn2020fixmatch} considers weakly-augmented samples as more reliable anchors and constrains its output to be the same as strongly-augmented data. Similarly, UDA~\cite{xie2020unsupervised} uses weakly-augmented data and complex-augmented data to generate similar output. CutMix~\cite{yun2019cutmix} is a widely adopted technique that generates pseudo labels and implicitly implements the idea of entropy minimization by ensuring the decision boundary passes through a low-density region of the distribution. CCT~\cite{ouali2020semi} and GCT~\cite{ke2020guided} use a similar idea to achieve feature augmentation through cross-confidence consistency. Consistency-based SSL methods with auxiliary networks can be considered network-level augmentation. In this paper, we propose applying complex augmentation (image-level augmentation and feature augmentation) to the students and weak augmentation to the teachers.
\section{Method}
We propose a novel consistency regularization framework based on mutual knowledge distillation, as described in Sec.~\ref{sec: mkd}. The image augmentation method is discussed in Sec.~\ref{sec: augmentation}. Finally, the training procedure is introduced in Sec.~\ref{sec: optimization}. We aim at training an end-to-end segmentation model with a massive amount of unlabeled and few labeled data in a semi-supervised learning manner.

\subsection{Mutual Knowledge Distillation Framework}
\label{sec: mkd}
\noindent \textbf{Overview.} 
We first present the settings for a typical semi-supervised semantic segmentation task. Labeled datasets and unlabeled datasets are denoted as $D^l=\{(\boldsymbol{x}_i^l, \boldsymbol{y}_i^l)\}_{i=1}^{|D^l|}$ and $D^u=\{(\boldsymbol{x}_i^u)\}_{i=1}^{|D^u|}$ with $|D^l| \ll |D^u|$, where $\boldsymbol{x}_i \in \mathbb{R}^{H \times W \times 3}$ is the input RGB image with the size of $H \times W$ and $\boldsymbol{y}_i \in \mathbb{R}^{H \times W \times C}$ represents the pixel-level one-hot label map for $C$ classes. The proposed MKD framework is illustrated in Figure~\ref{fig: framework}, which consists of four branches: two baseline student networks and two auxiliary mean teacher networks. The labeled images are fed into the student network and optimized with the normal cross-entropy loss $\mathcal{L}_{s}$ between ground truth labels. The unlabeled images with strong (weak) augmentation are fed into the student (teacher) networks. Each student network is trained under the supervision of the pseudo labels generated by the other student network ($\mathcal{L}_{ss}$) and by the other teacher network ($\mathcal{L}_{st}$). The knowledge between the two branches is transferred with the proposed MKD framework. Details for each part are described as follows.

\noindent \textbf{Baseline student networks.} 
The co-training methodology entails the establishment of two networks, each exhibiting identical structures but divergent initializations. This process is further characterized by the imposition of constraints to ensure consistency between the outputs of the two networks. Our baseline student networks are based on the previous state-of-the-art co-training method, CPS~\cite{chen2021semi}. Student networks are defined as $M_{s1}$ and $M_{s2}$, respectively. Network structures for $M_{s1}$ and $M_{s2}$ are the same, but the parameters are initialized differently with $\theta_{s1}$ and $\theta_{s2}$. For example, given an input image $\boldsymbol{x}$, the baseline student networks produce features denoted as $f_{s1}$ and $f_{s2}$, respectively. Following the typical co-training baseline~\cite{chen2021semi}, each student network is supervised by the pseudo labels generated by the other student network, which is denoted as  $\mathcal{L}_{ss}$.

\noindent \textbf{Auxiliary mean teacher networks.} Prior research, as demonstrated in~\cite{tarvainen2017mean, he2020momentum, grill2020bootstrap}, has illustrated the potential of mean teacher models to store and leverage historical information, thereby enhancing model performance. It does not need to be optimized, thus adding relatively little computation. Building on this idea, we incorporate two auxiliary mean teacher networks denoted as $M_{t1}$ and $M_{t2}$ into our MKD framework. The network structure of the teacher is the same as the network structure of the student. However, the mean teacher does not require back-propagation during training. The corresponding student model updates the mean teacher parameters according to exponential moving average(EMA) as Eq.~\eqref{eq: update teacher}, where $\gamma$ controls the speed of updates and $i \in [1,2]$ represents the index of the branch being updated:

\begin{align}
    \theta_{ti} = \gamma \theta_{ti} + (1-\gamma) \theta_{si}.
    \label{eq: update teacher}
\end{align}

\noindent \textbf{Mutual knowledge distillation.}
As illustrated in Figure~\ref{fig: framework}, labeled samples are used to train student models, and losses are calculated using supervised loss. Unlabeled samples, after strong augmentation, are fed into the two student models to obtain different outputs ($p_{s1}^u, p_{s2}^u$). Similarly, to generate more reliable supervision, samples after weak augmentation are fed into the two teacher models to obtain different outputs ($p_{t1}^u, p_{t2}^u$). There are two main objectives for the proposed MKD framework. First, we enable the teacher network to update smoothly and produce high-confidence predictions on easy samples. Second, we allow the student network to learn from more challenging samples and get more useful information. Thus, we apply all levels of augmentation to the samples fed into the student network. By adding two auxiliary mean teacher networks, we can obtain reliable supervision, which is not easy to collapse. To apply the network augmentation, the output pseudo label $\mathbf{y}_{t1}^{u}$ from the $M_{t1}$ is used to supervise the logits map $\mathbf{p}_{s2}^{u}$ from  $M_{s2}$, and vice versa. Eq.~\eqref{eq: cross-loss} is the consistency loss between teacher and student models:

\begin{align}
    \label{eq: cross-loss}
    \mathcal{L}_{st}^u=
    \ell_{st}^{\theta_{1}}\left (\mathbf{p}_{s1}^u, \mathbf{y}_{t2}^u \right) + 
    \ell_{st}^{\theta_{2}}\left (\mathbf{p}_{s2}^u, \mathbf{y}_{t1}^u \right ).
\end{align}

where $\mathbf{y}_{t1}^u$ and $\mathbf{y}_{t2}^u$ denotes one hot labels of the teachers' outputs.

\noindent \textbf{Knowledge selection.}\label{sec: knowledge selection}
We add a threshold on the teacher branch to confirm the training process supervised under higher confidence. This process is called knowledge selection, where the threshold ensures that the teacher is confident about transferring knowledge to students during the distillation process. By selecting a threshold greater than $0.95$, we leave noisy signals out and keep the supervision between student and teacher reliable. And we found that if the threshold is applied to the supervision between the student and the other student, some useful information is lost in this process, and worse results will be obtained. Following Eq.~\eqref{eq: unsupervised threshold}, we add a threshold to achieve the knowledge selection. 
$T$ is set to be $0.95$ by default in Eq. \eqref{eq: unsupervised threshold}:

\begin{align}
    \label{eq: unsupervised threshold}
    \mathcal{L}_{st}^u=
    \ell_{st}^{\theta_{1}}\left (\mathbf{p}_{s1}^u, \mathbf{y}_{t2}^u |(\mathbf{p}_{s2}^u \geq T) \right ) + 
    \ell_{st}^{\theta_{2}}\left (\mathbf{p}_{s2}^u, \mathbf{y}_{t1}^u |(\mathbf{p}_{s1}^u \geq T) \right ).
\end{align}

\subsection{Augmentation}
\label{sec: augmentation}
Pseudo-labels are generated by the model itself and provide limited information. To increase the diversity of the samples for the student network in our MKD framework, we apply image-level augmentations.

\noindent \textbf{Image augmentation.}
Image augmentation is based on weak and strong augmentation pairs. Weak data augmentation (WDA) (e.g.\  image flipping, cropping, resizing) is applied to the images passed to the teacher models. In addition, strong data augmentation (SDA) (e.g.\  image flipping, cropping, resizing, cutMix, random select an operator from color jitter, blur, gray-scale, equalize and solarize) is applied to the same ones fed to the student model to improve overall generalization. Motivated by the distribution of batch normalization~\cite{yuan2021simple}, we do not consider many strong color augmentation operations. Particularly, the CutMix~\cite{yun2019cutmix} augmentation is achieved by applying a binary mask $m$ that combines two images using the function $x = (1-m)\odot x_{i} + m\odot x_{j} \label{eq: cutmix_images}$. 
We apply CutMix by combining two input images in the batch for student models and use the same binary mask on the feature of teachers' logits with $p = (1-m)\odot p_{i} + m\odot p_{j} \label{eq: cutmix}$. Then we apply $p$ to supervising students.

\noindent \textbf{Feature augmentation.}
Feature Data Augmentation (FDA) employs limitless meaningful semantic transformations to modify the feature spaces. This method tweaks the image semantically without the need for an auxiliary network. FDA works by finding suitable translation vectors in the feature space and generating an enhanced feature set. For this enhancement process, category information is essential. However, for unlabeled data where such information is absent, we use pseudo-labels as a replacement.

When applying FDA to semi-supervised semantic segmentation, we meticulously augment each feature to develop an improved feature set. The network's refinement is achieved by reducing the cross-entropy (CE) loss. If we consider M tending towards infinity, we can compute the CE loss for all feasible augmented features. The upper bound of this loss is presented in Eq. \eqref{eq: ISDA}.

\begin{equation}
\label{eq: ISDA}
    \begin{aligned}
    \mathcal{L}_{\infty}(\theta \mid \boldsymbol{\Sigma})
     \leq \frac{1}{N} \sum_{i=1}^{N} \log (\sum_{j=1}^{C} e^{\boldsymbol{v}_{j y_{i}}^{\mathrm{T}} {\boldsymbol{f}}_{i} + (b_{j}-b_{y_{i}})+\frac{\lambda}{2} \boldsymbol{v}_{j y_{i}}^{\mathrm{T}} \Sigma_{y_{i}} \boldsymbol{v}_{j y_{i}}}).
    \end{aligned} 
\end{equation}

To incorporate semantic augmentation for semi-supervised semantic segmentation, we analyze the structure of the ISDA loss. Eq. \eqref{eq: ISDA} ends up with just one more term than the standard cross-entropy loss. Thus, each pixel in the student's features can be calculated by \label{eq: isda feature}$\mathbf{p}_{s} = {\boldsymbol{v}^{\mathrm{T}} {\boldsymbol{f}}_{s} + b + \frac{\lambda}{2} \boldsymbol{v}^{\mathrm{T}} \Sigma \boldsymbol{v}}$. Pseudo-labels are used to obtain the category information required for feature augmentation. Further details can be found in the supplementary materials.

\subsection{Optimization of the Framework}
\label{sec: optimization}
The full training loss for the whole framework is described in Eq. \eqref{eq: whole-loss}, where $\alpha$ and $\beta$ are the loss weights.
\vspace{-0.2em}
\begin{align}
    \label{eq: whole-loss}
    \mathcal{L} = 
    \mathcal{L}_{s}^l
    + \alpha \mathcal{L}_{st}^u 
    + \beta \mathcal{L}_{ss}^u,
\end{align}

The first loss in Eq. \eqref{eq: whole-loss} is the supervised segmentation loss for student models, defined as Eq. \eqref{eq: sup-loss}, 
where $\ell_{ce}$ is the cross-entropy loss function, $\mathbf{y}$ presents the ground truth,
and $\theta_{1}$ and $\theta_{2}$ are model parameters of different students.
\vspace{-0.2em}
\begin{align}
    \label{eq: sup-loss}
    \mathcal{L}_s^l=
    \ell_{ce}^{\theta_{1}}\left (\mathbf{p}_{s1}^l, \mathbf{y}^l \right) + 
    \ell_{ce}^{\theta_{2}}\left (\mathbf{p}_{s2}^l, \mathbf{y}^l \right ),
\end{align}

The second term is described in Eq. \eqref{eq: cross-loss}, which is the consistency loss based on cross-entropy. The last term $\ell_{ss}^u$ in Eq. \eqref{eq: whole-loss} is the consistency loss between students same as CPS~\cite{chen2021semi}. 
\vspace{-0.2em}
\begin{align}
    \label{eq: ss-loss}
    \mathcal{L}_{ss}^u=
    \ell_{ss}^{\theta_{1}}\left (\mathbf{p}_{s1}^u, \mathbf{y}_{s2}^u \right) + 
    \ell_{ss}^{\theta_{2}}\left (\mathbf{p}_{s2}^u, \mathbf{y}_{s1}^u \right ).
\end{align}

We show our MKD framework in Algorithm~\ref{algorithm}. First, we initialize student models with different random initialization parameters and set the same parameters for its teacher network. Then, after obtaining augmented data from the input images with SDA and WDA, we first use EMA to update the teachers' parameters. Finally, as described in Fig.~\ref{fig: framework}, we follow Eq.~\eqref{eq: whole-loss} to train the model.

\begin{algorithm}[htb]
\caption{\small Mutual Knowledge Distillation}
\label{algorithm} 
\footnotesize 
Initialize $\mathcal{L} \leftarrow 0$\ .\\
Sample labeled images $x^l$ and corresponding labels $y^l$ from $D^l$.\\

Sample unlabeled images without labels ${x}^u$ from $D^u$.\\
\textbf{Initialization Student} Randomly initialize two student models.\\
\textbf{Initialization Teacher} Apply the same initialization of each student to the corresponding teacher.\\
\For{${\rm step} = 1 ,..., n$ }{
    Update momentum of $M_{t1}$, $M_{t2}$. \\
    
    Apply weak data augmentation on ${x_{w}^{l}}$, ${x_{w}^{u}}$. \\
    Obtain strongly augmented ${x_{s}^{u}}$ 
    as described in 
    Sec.~\ref{sec: augmentation}. \\
    Generate $p^{u}_{t1}$, $p^{u}_{t2}$ as unlabeled probabilities.
    
    Gain labeled probabilities: $p^{l}_{s1}$, $p^{l}_{s2}$.

    Get featured augmented $p^{u}_{s1}$, $p^{u}_{s2}$
    based on Eq.~\eqref{eq: isda feature}. \\

    Calculate the whole loss as described in 
    Eq.~\eqref{eq: cross-loss}, \eqref{eq: whole-loss}, 
    \eqref{eq: sup-loss}, 
    \eqref{eq: ss-loss}.
}
\end{algorithm}
\section{Experiments}
\begin{table*}[t]
\caption{Compared with state-of-the-art methods on the Pascal VOC 2012 val set under different partition protocols. Here `1/n' means that we use `1/n' labeled dataset and the remaining images in the training set are used as the unlabeled dataset. $\dag$ means we introduce the unlabeled dataset with a total of 10582 images. $*$ denotes an enhanced training scheme, which will be further discussed in Table~\ref{tab: Impact of Each Module}. SupOnly stands for supervised training without using any unlabeled data. Blue text indicates the performance between our methods compared with the supervised-only method.}

\label{tab: SOTA on Pascal voc}
\begin{center}
\resizebox{\linewidth}{!}{
\setlength{\tabcolsep}{13pt}
\vspace{-5pt}
\begin{tabular}{l | lllll}
\toprule
Method  &1/16 (92) &1/8 (183) &1/4 (366) &1/2 (732) &Full (1464) \\
\midrule
\midrule
SupOnly-R50 & 43.97 & 49.57 & 57.76 & 65.73 & 68.81 \\
SupOnly-R101 & 45.77 & 54.92 & 65.88 & 71.69 & 72.50 \\
\midrule
CCT-R50~\cite{ouali2020semi}               & 33.10 & 47.60 & 58.80 & 62.10 & 69.40 \\
MT-R101~\cite{tarvainen2017mean}            & 48.70 & 55.81 & 63.01 & 69.16 & - \\
AdvSemSeg-R101~\cite{hung2018adversarial}   & 39.69 & 47.58 & 59.97 & 65.27 & 68.40 \\
VAT-R101~\cite{miyato2018virtual}           & 36.92 & 49.35 & 56.88 & 63.34 & - \\
CutMix-Seg-R101~\cite{yun2019cutmix}           & 55.58 & 63.20 & 68.36 & 69.84 & - \\ 
PC$^2$Seg-R101~\cite{zhong2021pixel}         & 57.00 & 66.28 & 69.78 & 73.05 & 74.15 \\
GCT-R101~\cite{ke2020guided}                & 46.04 & 54.98 & 64.71 & 70.67 & - \\
PseudoSeg-R101~\cite{zou2020pseudoseg}       & 57.60 & 65.50 & 69.14 & 72.41 & 73.23 \\
CPS-R101~\cite{chen2021semi}                 & 64.07 & 67.42 & 71.71 & 75.88 & - \\
SimpleBaseline-R101~\cite{yuan2021simple}    & -     & -     & -     & -     & 75.00 \\
\midrule
Ours-R50                                & 60.60 & 66.74 & 71.01 & 72.73 & 78.14 \\
Ours-R101                               & \textbf{65.35}
\footnotesize{(\textcolor{blue}{$+$19.58})} &
\textbf{70.18}
\footnotesize{(\textcolor{blue}{$+$15.26})}& 
\textbf{74.44} 
\footnotesize{(\textcolor{blue}{$+$8.56})}& 
\textbf{75.90} 
\footnotesize{(\textcolor{blue}{$+$4.21)}}& 
\textbf{79.96} 
\footnotesize{(\textcolor{blue}{$+$7.46})} \\

\midrule
\midrule

PS-MT-R101~\cite{liu2021perturbed}$^\dag$                        & 65.80 & 69.58 & 76.57 &78.42 & 80.01 \\
ST++-R101~\cite{yang2021st++}$^\dag$                & 65.20 & 71.00 & 74.60 & 77.30 & 79.10 \\
U$^2$PL-R101~\cite{wang2022semi}$^\dag$             & 67.98 & 69.15 & 73.66 & 76.16 & 79.49 \\

Ours-R101$^\dag$                        & 69.10 & 74.63 & 76.76 & 78.66 & 80.02 \\
Ours-R101$^\dag*$ & \textbf{76.12} \footnotesize{(\textcolor{blue}{$+30.35$})}
& \textbf{77.83} \footnotesize{(\textcolor{blue}{$+22.91$})}
& \textbf{80.40} \footnotesize{(\textcolor{blue}{$+14.52$})}
& \textbf{82.13} \footnotesize{(\textcolor{blue}{$+10.44$})}
& \textbf{83.78} \footnotesize{(\textcolor{blue}{$+11.28$})}\\
\midrule
\bottomrule
\end{tabular} 
}
\end{center}
\vspace{-1em}
\end{table*}

\subsection{Implementation Details}
\noindent \textbf{Datasets.}
Following previous methods~\cite{chen2021semi, yuan2021simple, wang2022semi}, experiments are performed on three widely used image segmentation datasets, PASCAL VOC 2012 (VOC)~\cite{everingham2015pascal}, Cityscapes~\cite{cordts2016cityscapes} and COCO~\cite{lin2014microsoft}. VOC~\cite{everingham2015pascal} is a standard semantic segmentation benchmark with 21 classes, including the background. The standard VOC datasets have 1464 images for training, 1449 images for validation, and 1456 images for testing. Following the previous works~\cite{chen2018encoder}, we combine VOC with 9118 training images from the Segmentation Boundary Dataset (SBD)~\cite{hariharan2011semantic} as VOCAug. During training on VOCAug and VOC, we employ a crop size of $512 \times 512$. 

Cityscpaes~\cite{cordts2016cityscapes} consists of 2975/500/1525 finely annotated urban scene images with resolution $2048 \times 1024$ for train/validation/test, respectively. The segmentation performance is evaluated over 19 challenging categories. We use a training crop size of $1024\times512$.

We have chosen the COCO dataset~\cite{lin2014microsoft} for our experiments to conduct further benchmarking of the proposed method. It is a challenging benchmark for semantic segmentation composed of 118k/5k for training/validation. We employ a crop size of $512 \times 512$.

\noindent \textbf{Training.} Our method is implemented on MMSegmentation~\cite{mmseg2020}. Following DeepLabV3Plus~\cite{chen2018encoder}, we use the ``poly" learning rate policy where the initial learning rate is multiplied by $(1 - iter/iter_{max})^{0.9}$. For VOC and COCO datasets, the initial learning rate is set to 0.0025, while for Cityscapes, it is set to 0.01. Specifically, the batch size is set to 16 for all datasets, and all training was performed on the four NVIDIA A100. We train the network with mini-bath stochastic gradient descent (SGD). The momentum is fixed as 0.9, and the weight decay is set to 0.0005. 

\noindent \textbf{Network architecture.}
We use DeepLabv3plus~\cite{chen2018encoder} with ResNet~\cite{he2016deep} pre-trained on ImageNet~\cite{krizhevsky2012imagenet} as our segmentation network for VOC and Cityscapes datasets. The decoder head is composed of separable convolution same as standard DeepLabv3plus. It is worth noting that we do not use any tricks in the model structure. We adopt Xception-65~\cite{chollet2017xception} as our backbone network for the COCO datasets, following the same architectural design as other methods for a fair comparison.

\noindent \textbf{Evaluation metrics.} Following~\cite{chen2018encoder}, we adopt the mean Intersection over Union (mIoU) as the evaluation metrics. All results are estimated on the validation set. Particularly, we report results via only single-scale testing.

\vspace{-1em}
\subsection{Comparison with State-of-the-art Methods}
\begin{table*}[htpb]
\caption{Comparison with state-of-the-art on the PASCAL VOCAug and Cityscapes val set under different partition protocols. The VOCAug trainset consists of 10,582 labeled samples in total. $\ddag$ means the same split as U$^2$PL. Other methods use the same split as CPS. $^\bigstar$ presents the approach reproduced by~\cite{wang2022semi}. (-) means data is not available. The underline represents the best result of the CPS split. The best results of the U$^2$PL split are shown in bold. SupOnly stands for supervised training without using any unlabeled data. Blue text indicates the performance between our methods compared with the supervised-only method.}
  \label{tab: compared with sota}
  \centering
  \resizebox{\linewidth}{!}{
  \begin{tabular}{@{}l| llllllll@{}}
  \toprule
  \multicolumn{1}{c}{Method} & 
  \multicolumn{4}{c}{ResNet-50} & 
  \multicolumn{4}{c}{ResNet-101} \\ 
  \cmidrule(lr){2-5} \cmidrule(lr){6-9}
  \multicolumn{1}{c}{} &
  \multicolumn{1}{c}{1/16(662)} &
  \multicolumn{1}{c}{1/8(1323)} &
  \multicolumn{1}{c}{1/4(2646)} &
  \multicolumn{1}{c}{1/2(5291)} &
  \multicolumn{1}{c}{1/16(662)} &
  \multicolumn{1}{c}{1/8(1323)} &
  \multicolumn{1}{c}{1/4(2646)} &
  \multicolumn{1}{c}{1/2(5291)}  \\ \midrule
  \multicolumn{9}{c}{\emph{Pascal VOC 2012}} \\ \midrule
  SupOnly & 62.40 & 68.20 & 72.30 & - & 70.60 & 73.12 & 76.35 & 77.21 \\
   SupOnly$^\ddag$ & - & - & - & - & 67.87 & 71.55 & 75.80 & 77.13 \\
  \hline
  MT~\cite{tarvainen2017mean} 
  & 66.77 & 70.78 & 73.22 & 75.41 
  & 70.59 & 73.20 & 76.62 & 77.61 \\
  CCT~\cite{ouali2020semi} 
  & 65.22 & 70.87 & 73.43 & 74.75 
  & 67.94 & 73.00 & 76.17 & 77.56 \\
  CutMix-Seg~\cite{yun2019cutmix} 
  & 68.90 & 70.70 & 72.46 & 74.49 
  & 72.56 & 72.69 & 74.25 & 75.89 \\
  GCT~\cite{ke2020guided} 
  & 64.05 & 70.47 & 73.45 & 75.20 
  & 69.77 & 73.30 & 75.25 & 77.14 \\
  CPS~\cite{chen2021semi} 
  & 68.21 & 73.20 & 74.24 & 75.91 
  & 72.18 & 75.83 & 77.55 & 78.64 \\
  CPS w/ CutMix~\cite{chen2021semi} & 71.98 & 73.67 & 74.90 & 76.15 & 74.48 & 76.44 & 77.68 & 78.64 \\
  PS-MT~\cite{liu2021perturbed}
  & 72.83 & 75.70 & 76.43 & 77.88
  & 75.50 & 78.20 & 78.72 & 79.76 \\
  ST++~\cite{yang2021st++}
  & 73.20 & 75.50 & 76.00 & - 
  & 74.70 & 77.90 & 77.90 & -  \\
  U$^2$PL$^\ddag$~\cite{wang2022semi}   
  & - & - & - & - 
  & 77.21 & 79.01 & 79.30 & 80.50 \\
  \hline
  Ours & 
  \underline{76.88} \footnotesize{(\textcolor{blue}{$+$14.48})} & 
  \underline{78.25}
  \footnotesize{(\textcolor{blue}{$+$10.05})} & 
  \underline{79.23}
  \footnotesize{(\textcolor{blue}{$+$6.93})} & 
  \underline{79.74}
  \footnotesize{(\textcolor{blue}{-})} 

  & \underline{78.65} \footnotesize{(\textcolor{blue}{$+8.05$})} & \underline{80.11} \footnotesize{(\textcolor{blue}{$+6.99$})} & \underline{80.75} \footnotesize{(\textcolor{blue}{$+4.40$})} & \underline{81.72} \footnotesize{(\textcolor{blue}{$+4.51$})} \\

  Ours$^\ddag$ & 
  \textbf{77.99}
  \footnotesize{(\textcolor{blue}{-})}& 
  \textbf{78.49}
  \footnotesize{(\textcolor{blue}{-})}& 
  \textbf{78.86}
  \footnotesize{(\textcolor{blue}{-})}& 
  \textbf{78.19}
  \footnotesize{(\textcolor{blue}{-})}& 
  \textbf{80.05}
  \footnotesize{(\textcolor{blue}{$+$12.18})}& \textbf{81.35}
  \footnotesize{(\textcolor{blue}{$+$9.80})}& \textbf{82.30}
  \footnotesize{(\textcolor{blue}{$+$6.50})}& \textbf{80.60} \footnotesize{(\textcolor{blue}{$+$3.47})}\\

  \midrule
  & \multicolumn{4}{c}{ResNet-50} & \multicolumn{4}{c}{ResNet-101} \\ 
  \cmidrule(lr){2-5} \cmidrule(lr){6-9}
  \multicolumn{1}{c}{} &
  \multicolumn{1}{c}{1/16(186)} &
  \multicolumn{1}{c}{1/8(372)} &
  \multicolumn{1}{c}{1/4(744)} &
  \multicolumn{1}{c}{1/2(1488)} &
  \multicolumn{1}{c}{1/16(186)} &
  \multicolumn{1}{c}{1/8(372)} &
  \multicolumn{1}{c}{1/4(744)} &
  \multicolumn{1}{c}{1/2(1488)}  \\ 
  \midrule
  \multicolumn{9}{c}{\emph{Cityscapes}} \\ \midrule
  SupOnly  & - & 70.80 & 73.70 & - & 62.96 & 69.81 & 74.23 & 77.46 \\
  SupOnly$^\ddag$ & - & - & - & - & 65.74 & 72.53 & 74.43 & 77.83 \\
  \midrule
  MT~\cite{tarvainen2017mean} 
  & 66.14 & 72.03 & 74.47 & 77.43 
  & 68.08 & 73.71 & 76.53 & 78.59 \\
  CCT~\cite{ouali2020semi} 
  & 66.35 & 72.46 & 75.68 & 76.78 
  & 69.64 & 74.48 & 76.35 & 78.29 \\
  CutMix-Seg~\cite{french2019semi} 
  & - & - & - & - 
  & 72.13 & 75.83 & 77.24 & 78.95 \\
  GCT~\cite{ke2020guided} 
  & 65.81 & 71.33 & 75.30 & 77.09 
  & 66.90 & 72.96 & 76.45 & 78.58 \\
  CPS~\cite{chen2021semi} 
  & 69.79 & 74.39 & 76.85 & 78.64 
  & 70.50 & 75.71 & 77.41 & 80.08 \\
  CPS w/ CutMix~\cite{chen2021semi} & \underline{74.47} & 76.61 & 77.83 & 78.77 & 74.72 & \underline{77.62} & \underline{79.21} & 80.21 \\
  CPS$^{\ddag}{^\bigstar}$~\cite{chen2021semi} 
  & - & - & - & - 
  & 69.78 &  74.31 &74.58 & 76.81 \\
  SimpleBaseline~\cite{yuan2021simple} 
  & - & - & - & - 
  & - & 74.10 & 77.80 & 78.70 \\
  PS-MT~\cite{liu2021perturbed}
  & - & \underline{77.12} & 78.38 & 79.22 
  & - & - & - & - \\
  U$^2$PL$^\ddag$~\cite{wang2022semi} 
  & - & - & - & - 
  & 70.30 & 74.37 & 76.47 & 79.05 \\
  
  \hline
  Ours 
  & \textbf{75.47}
  \footnotesize{(\textcolor{blue}{-})}
  & \textbf{78.07}
  \footnotesize{(\textcolor{blue}{$+$7.27})}
  & \textbf{79.95}
  \footnotesize{(\textcolor{blue}{$+$6.25})}
  & \textbf{80.52} \footnotesize{(\textcolor{blue}{-})}
  & \textbf{77.19}
  \footnotesize{(\textcolor{blue}{$+$14.23})}
  & \textbf{79.20}
  \footnotesize{(\textcolor{blue}{$+$9.39})}
  & \textbf{80.80}
  \footnotesize{(\textcolor{blue}{$+$6.57})}
  & \textbf{81.04}
  \footnotesize{(\textcolor{blue}{$+$3.58})}\\
  Ours$^\ddag$ 
  & 73.14 
  \footnotesize{(\textcolor{blue}{-})}& 
  75.00 
  \footnotesize{(\textcolor{blue}{-})}& \underline{78.62} \footnotesize{(\textcolor{blue}{-})}
  & \underline{79.90} \footnotesize{(\textcolor{blue}{-})}
  & \underline{75.31}
  \footnotesize{(\textcolor{blue}{$+$9.57})}
  & 75.98 
  \footnotesize{(\textcolor{blue}{$+$3.45})}
  & 78.28 
  \footnotesize{(\textcolor{blue}{$+$3.85})}& \underline{80.74}
  \footnotesize{(\textcolor{blue}{$+$2.91})} \\
  \bottomrule
  \end{tabular}
  }

\end{table*}

We conduct the comparison experiments with state-of-the-art algorithms in Table~\ref{tab: SOTA on Pascal voc}, Table~\ref{tab: compared with sota}, and Table~\ref{tab: SOTA COCO}.

\begin{table*}[htpb]
\caption{
Comparison with state-of-the-art on the COCO~\cite{lin2014microsoft} dataset based on Xception-65~\cite{chollet2017xception} under different partition protocols. SupOnly stands for supervised training without using any unlabeled data. Blue text indicates the performance between our methods compared with the supervised-only method.}
\label{tab: SOTA COCO}
\begin{center}
\resizebox{0.7\linewidth}{!}{
\begin{tabular}{l | l l l l l}
\toprule
Method & 1/512(232) & 1/256(463) & 1/128(925) & 1/64(1849) &1/32(3697)) \\
\midrule
SupOnly & 22.9 & 28.0 & 33.6 & 37.8 & 42.2 \\
\midrule

PseudoSeg~\cite{zou2020pseudoseg} &29.8 &37.1 &39.1 &41.8 &43.6\\
PC$^2$Seg~\cite{zhong2021pixel} &29.9 &37.5 &40.1 &43.7 &46.1\\
Ours      
&\textbf{36.7} \footnotesize{(\textcolor{blue}{$+$13.8})} &\textbf{43.7}
\footnotesize{(\textcolor{blue}{$+$15.7})}
&\textbf{48.9}
\footnotesize{(\textcolor{blue}{$+$15.3})}
&\textbf{51.0}
\footnotesize{(\textcolor{blue}{$+$13.2})}
&\textbf{54.1}
\footnotesize{(\textcolor{blue}{$+$11.9})}\\
\bottomrule
\end{tabular}
}

\end{center}
\vspace{-1em}
\end{table*}

\begin{figure*}[htpb]
    \centering
    \includegraphics[width=\linewidth]{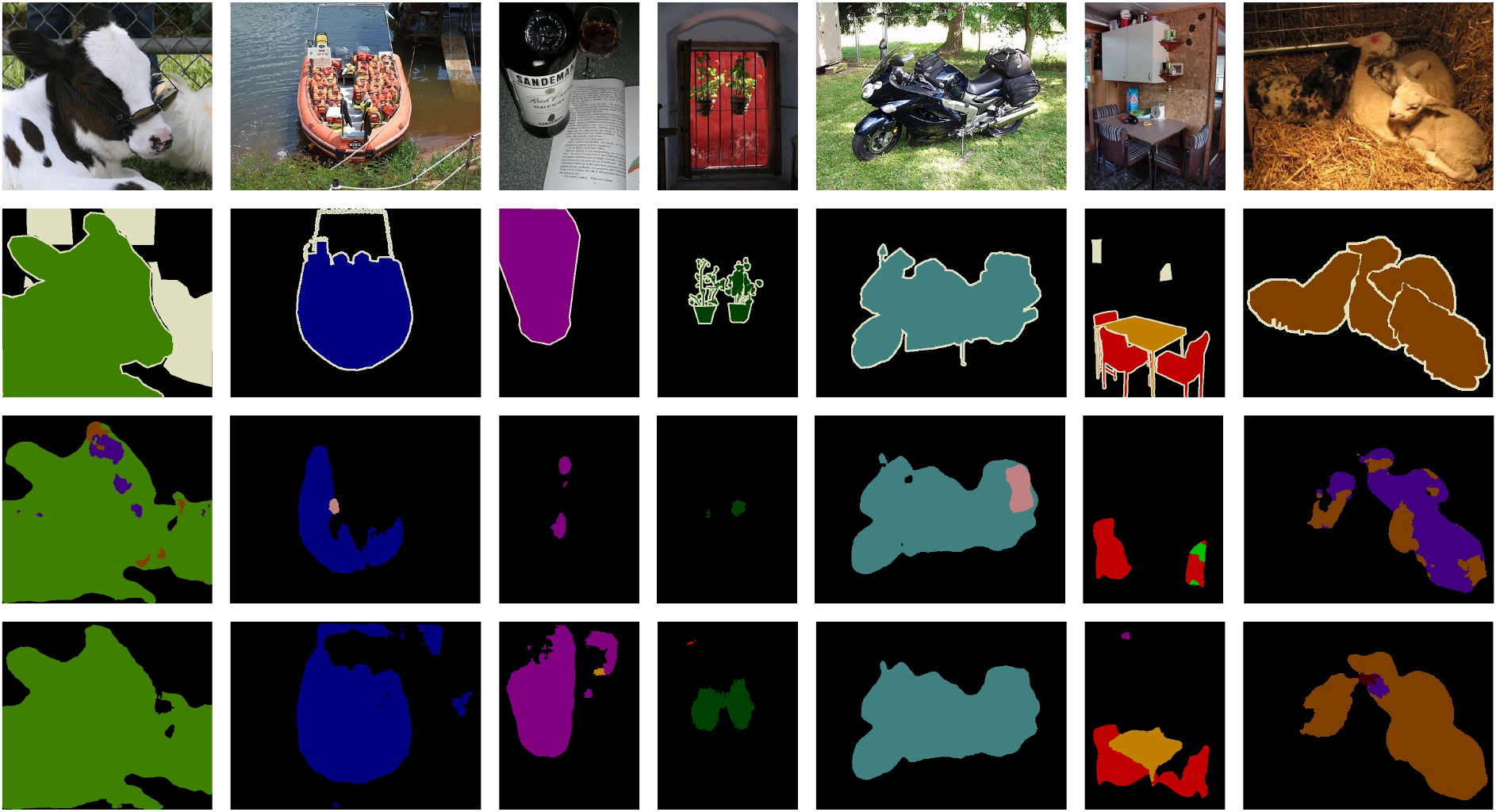}
    \caption{Qualitative results on the PASCAL VOC 2012 datasets using 1/16 (662) labeled samples and ResNet50.
    The first line exhibits the input images. The second line shows the ground truth. The third line presents the baseline of co-training. The fourth line displays our method.}
    \label{fig: visual voc}
    \vspace{-0.51em}
\end{figure*}

\noindent\textbf{Results on PASCAL VOC 2012 dataset.}
Table~\ref{tab: SOTA on Pascal voc} shows comparison results on PASCAL VOC 2012 dataset. Following previous settings, we sample labeled images from 1) the original VOC 1464 training images, and 2) VOCAug with a total of 10582 images. It is important to notice that the methods with $\dag$ and without $\dag$ only differ in unlabeled images. They share the same `1/n' labeled data set and validation set with 1449 images. On the first data splits, assuming 1464 images in total for training, the proposed framework accomplishes 1.28\%, 2.76\%, 2.73\%, with only 92, 183, and 366 labeled images under ResNet-R101 compared with CPS-R101~\cite{chen2021semi}.  

Note that we achieve similar performance with CPS~\cite{chen2021semi} under 1/16 partitions, which is trained with only 92 label images as the number of labeled images is too small to generate reliable labels for the teacher network. 
As shown in Table~\ref{tab: SOTA on Pascal voc}, our method improved by 30.35\% based on ResNet-101 compared with the supervised-only method with only 92 labeled images on VOCAug. When additional unlabeled data are incorporated, we adopt the identical splits used in previous works such as~\cite{yang2021st++, wang2022semi, liu2021perturbed}.

Our method still gains remarkable performance based on ResNet-101. It shows that more unlabeled images could bootstrap performance. In particular, compared with the U$^2$PL method, our method improves by 8.14\% and 8.38\% under 1/16 and 1/8 partition protocols. It is also demonstrated that our method is more effective on fewer data. In addition, we set a confidence of 0.95 for selecting regions with higher confidence to select useful knowledge. The results are reported at the last line in table~\ref{tab: SOTA on Pascal voc}. 

Table~\ref{tab: compared with sota} compares our method with the other state-of-the-art methods on VOCAug. To make a fair comparison, we train our MKD framework under two different split lists following previous work. Using the same split as CPS, the proposed method performs favorably against the previous state-of-the-art methods. Furthermore, as the amount of data increases, the performance gap between the various methods becomes smaller, proving that the segmentation task does not require a lot of labeled data.

Figure~\ref{fig: visual voc} shows the quantitative results of different methods on the PASCAL VOC 2012 datasets. We can see that co-training can not reasonably separate the objects (especially large-sized objects such as cows, boats, sheep, and motorbikes) completely while ours corrects these errors. Compared to co-training, our method performs well on these complex examples, such as potted
plant and chair. 

\noindent\textbf{Results on Cityscapes dataset.} 
The Cityscapes dataset consists of images focusing on urban scenes. As shown in Table~\ref{tab: compared with sota}, our method achieves notable improvements under various partition protocols with the same split as CPS~\cite{chen2021semi}. In addition, we improve by 5.01\% under 1/16(186) partition protocol with the same split as U$^2$PL~\cite{wang2022semi}. Our method outperforms the existing state-of-the-art method by a notable margin. Specifically, we report results with single-scale testing. We attribute this significant improvement to the fact that the Cityscapes dataset is relatively redundant so the teacher model can provide more accurate pseudo-labeling.

\noindent\textbf{Results on the COCO dataset.} 
The COCO dataset is a quite challenging task with 118k training images, consisting of 81 classes in total.
As shown in Table~\ref{tab: SOTA COCO}, our method achieves much better results compared with PC$^2$Seg~\cite{zhong2021pixel} based Xception-65~\cite{chollet2017xception} under 1/512, 1/256, 1/128, 1/64 and 1/32 partition protocols the same as PseudoSeg~\cite{zou2020pseudoseg}. In addition, we improve by 6.2\%-8.8\% with the same split as PC$^2$Seg~\cite{zhong2021pixel}. Our method outperforms the existing state-of-the-art method by a notable margin.

\vspace{-1.2em}

\label{sec: ablation study}
\subsection{Ablation Study}

\begin{table*}[ht]
\caption{
Ablation study on the proposed semi-supervised learning framework. The model here is Deeplabv3Plus with ResNet101 backbone. Co-training denotes the baseline the same as CPS. Mutual MT presents mutual knowledge distillation. SDA denotes strong data augmentation. KS is knowledge selection. FDA denotes feature data augmentation}
\label{tab: Impact of Each Module}
\begin{center}
\resizebox{0.95\linewidth}{!}{

\begin{tabular}{c c c c c c c c c c c}
\hlineB{2}
Co-training & Mutual MT & SDA & KS & FDA & 1/16(662) & 1/8(1323) & 1/4(2646) & 1/2(5291)  \\
\hline

\checkmark & & & & & 72.18 & 75.83 & 77.55 & 78.64 \\
\checkmark  &\checkmark & & & & 75.19 & 77.92 & 79.76 & 81.03 \\
\checkmark  &\checkmark & & & \checkmark &  77.52 & 78.53  & 79.45 & 80.63   \\
\checkmark  & &\checkmark & & & 74.48 & 76.44 & 77.68 & 78.64 \\
\checkmark  &\checkmark &\checkmark & & & 78.00 & 79.49 & 80.66 & 80.77 \\

\checkmark  &\checkmark & \checkmark &\checkmark & & 78.65 & 80.11 & 80.75 & 81.72   \\
\hlineB{2}
\end{tabular}
}

\end{center}
\end{table*}

\begin{table*}[htbp]
\makeatletter\def\@captype{table}\makeatother

\caption{Ablation study of the knowledge selection with different components $\mathcal{L}_{ss}$ and $\mathcal{L}_{st}$. It uses R101 as the backbone with PseudoSeg splits.
}

\makeatletter\def\@captype{table}\makeatother

\vspace{-0.5ex}
\label{tab: Impact of threshold component}
\resizebox{0.65\linewidth}{!}{

\begin{tabular}{c c c c c c c }
\hlineB{1}
$\mathcal{L}_{st}$ & $\mathcal{L}_{ss}$ & 1/16(92) & 1/8(183) & 1/4(366) & 1/2(732) & Full(1464)\\
\hline
\checkmark &  &74.72 &77.16 &79.15 &80.45 &81.81\\
&\checkmark &73.20 &75.65 &78.42 &80.23 &82.39\\
\checkmark& \checkmark & 73.41 &75.64 &78.71 &80.30 &82.12\\
\hlineB{1}
\end{tabular}
}

\end{table*}

\begin{table*}[htbp]
\makeatletter\def\@captype{table}\makeatother

\caption{Ablation study of the heterogeneous network augmentation, which uses ResNet101 as the backbone. SN means the same network, and HN means heterogeneous network.
}

\makeatletter\def\@captype{table}\makeatother

\vspace{-0.5ex}
\label{tab: Impact of HN}
\resizebox{0.65\linewidth}{!}{

\begin{tabular}{c c c c c c c }
\hlineB{1}
HN & SN & 1/16(92) & 1/8(183) & 1/4(366) & 1/2(732) & Full(1464)\\
\hline
\checkmark &  &74.89 &77.63 &79.48 &81.13 & 82.54 \\
&\checkmark &76.12 &77.83 &80.40 &82.13 &83.78\\
\hlineB{1}
\end{tabular}
}

\end{table*}

In this subsection, we conduct experiments to explore the effectiveness of each proposed module on the VOC dataset under different semi-supervised settings.

\noindent \textbf{Effectiveness of mutual knowledge distillation.} 
As illustrated in Table \ref{tab: Impact of Each Module}, we conduct a series of experiments to identify each module's performance. We take co-training as our baseline, the same as CPS~\cite{chen2021semi}. We first try to add naive mean teacher (MT) and find that the results do not improve or reduce and even lead to training instability. It may be because the teacher and student models are too similar, leading to collapse. By adding the mutual mean teacher, the collapse disappeared, which resulted in a 3.01\% performance improvement under 1/16 (662).

\noindent \textbf{Effectiveness of data augmentation.} 
In Table \ref{tab: Impact of Each Module}, to introduce more augmentation, we also added the strategies of strong augmentation(SDA) which accompanied 2.30\% performance improvement under 1/16(662) partition protocols. Combining SDA and Mutual MT, we improve original co-training from 72.18\% to 78.00\% resulting in a 5.82\% gain. The final combination of all methods obtains the best result and yields a performance improvement of 6.47\% under 1/16(662) partition protocols.

\noindent\textbf{Ablation study on feature augmentation.} We conducted an ablation study on the effect of feature augmentation using the ResNet101 backbone on the CPS splits. Our results indicate that the proposed approach yields a performance boost of 2.33\% and 0.61\% for 1/16 and 1/8 labeled data ratios, respectively. However, we also observed that this approach can harm training performance when the labeled data is overwhelming. The details of this are discussed in the supplementary section.

\noindent \textbf{Ablation study on knowledge selection.} The Table \ref{tab: Impact of threshold component} shows the effectiveness of the knowledge selection with different components $\mathcal{L}_{ss}$ and $\mathcal{L}_{st}$. It can be seen that the knowledge selection applied on $\mathcal{L}_{st}$ is superior to others, indicating that the improvement brought by knowledge selection applied to students and teachers will leave noise out with lower confidence. And the simple form of knowledge selection with a 0.95 threshold is reliable, but other reasonable values are also acceptable.

\noindent \textbf{Ablation study on heterogeneous network augmentation.}
To assess the effectiveness of our method, we conducted an experiment comparing the same model architecture with the heterogeneous network (HN) models, which employed PSPNet and Deeplabv3plus as teachers. The results, shown in Table \ref{tab: Impact of HN}, indicate that our method achieved a 1.23\% improvement over the HN under the 1/16 (662) partition protocol, demonstrating the superiority of our method.
\section{Conclusion}
We have proposed a new consistency learning scheme, called mutual knowledge distillation, for semantic segmentation. Our method utilizes two auxiliary mean-teacher models and a combination of strong-weak augmentation and feature augmentation to increase the diversity of training samples for the student network. Experimental results show that our proposed method outperforms recent state-of-the-art methods on several benchmark datasets for semantic segmentation, including PASCAL VOC 2012, Cityscapes, and Microsoft COCO. Notably, our framework achieves significant performance improvements even when labeled data is limited.

{\small
\bibliographystyle{ACM-Reference-Format}
\bibliography{sample-base}
}

\end{document}